\documentclass[letterpaper]{article} % DO NOT CHANGE THIS
\usepackage{aaai2026}  % DO NOT CHANGE THIS
\usepackage{times}  % DO NOT CHANGE THIS
\usepackage{helvet}  % DO NOT CHANGE THIS
\usepackage{courier}  % DO NOT CHANGE THIS
\usepackage[hyphens]{url}  % DO NOT CHANGE THIS
\usepackage{graphicx} % DO NOT CHANGE THIS
\usepackage{natbib}  % DO NOT CHANGE THIS AND DO NOT ADD ANY OPTIONS TO IT
\usepackage{caption} % DO NOT CHANGE THIS AND DO NOT ADD ANY OPTIONS TO IT
\usepackage{longtable}
\usepackage{subcaption}
\usepackage{multirow}
\usepackage[table,xcdraw]{xcolor}

\title{Inductive Claims Extraction at Scale}
 
\author{
    Sandrine Chausson,
    Björn Ross
}
\affiliations{
    University of Edinburgh\\
    sandrine.chausson@ed.ac.uk, b.ross@ed.ac.uk
}

\begin{document}
 
\maketitle
 
\begin{abstract}
A large part of political discourse on social media is built and expressed at a level of claims: i.e. declarative, typically single-clause statements, which convey a particular interpretation of reality and can range from factual to evaluative. Moreover, rather than occurring randomly, claims coalesce, recur in patterns, and come to be associated with different world views. When paired with structural computational tools such as Social Network Analysis, claims can be a powerful unit of analysis to study political phenomena such as echo chambers or polarisation. In this paper, we present a pipeline that uses a large language model (LLM) to inductively extract and catalogue claims from large social media corpora, and apply it to two different Twitter datasets: one relating to the 2020 US presidential election and the other to the 2022 FIFA World Cup. We comprehensively evaluate the approach by measuring the pipeline's recall and precision against manually annotated samples, run ablation studies isolating the contribution of its various components, and perform a qualitative error analysis. We discuss the value of the approach in the context of Computational Social Science research, and illustrate its capabilities by presenting the claims catalogue obtained from each dataset.
\end{abstract}
 
% Uncomment once code/data/extended version are ready to link.
% \begin{links}
%     \link{Code}{https://github.com/ANONYMIZED/claims-extraction}
%     \link{Data}{https://ANONYMIZED}
% \end{links}
 
\section{Introduction}

Social media is one of the primary places where political world views are built, shared, and expressed. But what makes a world view recognizable, and distinguishes it from another, sits at a level that the tools most commonly used to study online discourse at scale are not built to capture. Topic modelling recovers what a corpus is \emph{about}: a cluster of posts about ``Covid'', ``LGBTQ rights" or ``Western media bias" identifies a shared discursive agenda, but says nothing about what its authors believe. This holds whether topics are recovered from word frequencies \cite{blei2003lda}, by clustering document embeddings \cite{grootendorst2022bertopic}, or by prompting an LLM to label those clusters \cite{pham2024topicgpt}. Stance and sentiment classification \cite{mohammad2016semeval, aldayel2021stance} recover a coarse orientation (for or against, positive or negative) toward a target the researcher has already specified, but collapse the actual content of that position into a single sign. Misinformation research \cite{vosoughi2018spread, lazer2018science} comes closer, in that it engages directly with the content of a worldview, but by design restricts itself to the subset of that content that is checkable against external fact. What all three leave out, in different ways, is the level of the \emph{claim}: a short, declarative statement that conveys a more or less objective interpretation of political reality, whether strictly verifiable (``the election was stolen'') or explicitly evaluative and therefore resistant to any fact-check (``This World Cup is great'') \cite{chausson_insight-inference_2026}. 

Some computational work already engages with the claim as a unit of analysis, but it does so in a confirmatory, deductive way: given a pre-specified claim, it asks whether or not that claim is present in a specific piece of text \cite{chausson_detecting_2024, chausson_insight-inference_2026}. Doing so though presupposes a taxonomy of the claims that matter in the corpus of interest. Taxonomies of this sort have traditionally been built by social scientists through manual hand-coding or interviews. While that qualitative work remains indispensable, it cannot always scale to corpora of the size social media now produces. What is missing is therefore an inductive computational approach to claims, one that constructs the taxonomy itself directly from the data. Current Large Language Models (LLMs) offer a natural way to meet this need: when deployed appropriately, existing research has shown they can often outperform crowdsourced annotators \cite{gilardi2023chatgpt, tornberg2023chatgpt, ziems2024can}.

Here, we present a pipeline for LLM-based claims extraction and apply it to two large Twitter corpora, relating to the 2020 US presidential election and the 2022 FIFA World Cup respectively. The paper makes four contributions:

\begin{itemize}
    \item The formalisation of \emph{inductive claims extraction} as a Computational Social Science task, distinct from topic modelling, stance detection and deductive claims detection, and a working definition of the claim as its unit of analysis.
    \item A pipeline for inductively extracting and cataloguing claims from large social media corpora using an LLM. % method contribution
    \item A comprehensive and fine-grained validation of the pipeline on two thematically distinct large-scale corpora (2020 US election and 2022 FIFA World Cup).
    \item Two public catalogues of claims derived from these corpora (594 claims from the US elections corpus, and 237 claims from the FIFA World Cup corpus), with information on how these are distributed across different retweet communities.
\end{itemize}

\section{Related Work}\label{sec:related_work}

\subsection{Defining claims}

The concept of ``claim" is most extensively developed in Argument Mining \cite{lippi2016argumentation, lawrence2020argument}, where, following the Toulmin model, it denotes the assertion a speaker puts forward for acceptance, stripped of the evidence offered in its support \cite{toulmin_uses_1958}, and is treated as the conclusion of an argument in persuasive discourse \cite{stab-gurevych-2017-parsing, daxenberger-etal-2017-essence}.

Social media discourse rarely fits this persuasive frame: a user posting ``Trump is great'' is plausibly expressing a position rather than arguing for it, and the claims of interest to computational social scientists are correspondingly closer to evaluative statements than to argumentative claims proper. A looser, standalone conception is already standard outside Argument Mining. Check-worthiness detection treats a claim as any factual assertion of public interest, regardless of argumentative structure \cite{hassan-etal-2017-toward}, and ``deductive" NLI-based claims detection tests each claim of a pre-specified taxonomy for entailment against a document, independent of whether that document is itself an argument \cite{chausson_detecting_2024, chausson_insight-inference_2026}.

In this paper, we adopt this broader conception of claim. We define a claim as a declarative statement, simple enough to be expressed as a sentence with a single clause, in line with \citet{chausson_insight-inference_2026}. Moreover, we require that this claim be evaluative or attributional: i.e. it must ``make a point", express an opinion, or take a stance. This requirement is what distinguishes claims like ``Trump is great" from more neutral ones like ``It's raining".

\subsection{Top-Down Claims Detection}

A substantial body of work addresses the ``top-down" problem of determining whether a claim is present in, or supported by, a given piece of text. Early systems such as ClaimBuster treat this problem as supervised classification, training a model directly on hand-labeled sentences to flag factual, check-worthy claims in political speech \cite{hassan-etal-2017-toward}. Fact-verification pipelines such as those built on the FEVER dataset instead decompose the problem into retrieving candidate evidence for a claim and then classifying the relationship between claim and evidence \cite{thorne2018fever}, typically using models trained for Natural Language Inference (NLI) \cite{williams2018broad} to determine whether the evidence supports, refutes, or is unrelated to the claim. A related but more lightweight use of NLI dispenses with the retrieval step altogether, applying an entailment model directly to a candidate claim and a document (e.g. a social media post) to determine whether the claim is present in it \cite{yin2019benchmarking, laurer2024less, chausson_detecting_2024, chausson_insight-inference_2026}.

What all of these methods share, and what distinguishes them from the present work, is that they presuppose the claim (or taxonomy of claims) is already known. Our approach is therefore best understood as complementary to, rather than competing with, these more ``top-down" methods: it generates the taxonomy of claims that these methods take as a given input.

\subsection{Bottom-Up Claims Extraction}

A smaller body of work addresses the harder problem this paper is concerned with: discovering which claims populate a corpus, without knowing in advance what they are. Early approaches to this problem were largely inherited from Open Information Extraction, which extracts subject-relation-object tuples from text without a fixed schema \cite{banko2007open, fader2011identifying}. Applied to claims, this yields bare propositions rather than claims that make a point in the sense described above: Open IE has no mechanism for distinguishing an evaluative or attributional statement from a neutral one. Later work adapted this open-extraction logic to political text specifically, mining fine-grained political statements from blog comments \cite{bamman2015open} or agent--action--patient narratives from congressional speech \cite{ash2024relatio}, but inherits the same limitation: what is extracted is determined by syntactic structure, not by whether a statement takes a side. Within Argument Mining specifically, unsupervised corpus-level approaches have been proposed as a way to surface claims for a given topic without a predefined taxonomy \cite{levy2017unsupervised}, but this line of work still operates within an argumentative framing, and remains topic-driven: the analyst must name the topic in advance (a single concept with a Wikipedia entry), and only sentences mentioning it in a fixed lexical pattern are retrieved.

More recently, LLMs have been shown to perform competitively on free-form coding and generation tasks, in some cases producing explanations judged comparable to or better than crowdworker references \cite{ziems2024can}. Closest to our task, \citet{hoyle2023decompositions} prompt an LLM to decompose political texts into sets of propositions, including ones only implicitly conveyed, and show that these improve downstream text representations. While this evidence concerns open-ended coding and explanation generation rather than corpus-level claim extraction specifically, it is suggestive of LLMs' suitability for the same kind of inductive, corpus-grounded labelling tasks as claim extraction. 

Our own pipeline itself rests on an LLM-prompting strategy and therefore sits within this last line of work. However, it extends it in two ways. First, our focus on recurring, evaluative claims at the corpus level represents a narrower and comparatively unexplored application of LLM-prompting in Computational Social Science. Second, it is designed to approximate a corpus far larger than an LLM could process directly --- 600K and 574K tweets in total in each of our worked examples --- by combining LLM-prompted extraction with the stratified sampling logic long used in qualitative hand-coding approaches \cite{krippendorff2018content, nelson2020computational}.

\section{Data}
\label{sec:data}

We illustrate and evaluate our proposed claims extraction methodology on two large Twitter corpora from the literature. These relate to the 2020 US presidential elections \cite{chausson_insight-inference_2026} and 2022 FIFA World Cup \cite{chausson_beyond_2026} respectively. 

Both corpora have clear yet complementary relevance to the study of political discourse. The 2020 US presidential election is overtly and continuously political: its discourse is explicitly organized around candidates, parties, and an electoral outcome. The 2022 FIFA World Cup, by contrast, is not explicitly political in this sense: its discourse is nominally about a sporting competition. Yet it has been shown to carry a substantial undercurrent of genuinely political content, i.e. ``political talk", embedded within what otherwise reads as match commentary and fandom activity \cite{chausson_beyond_2026}. 

Both datasets are also Twitter corpora specifically, meaning that they contain a structural dimension in the form of retweets. Retweet networks are a well-established basis for structural community detection \cite{conover-etal-2011-political, barbera2015tweeting, cinelli2021echo}. In this paper, we therefore use retweet networks to identify sub-networks of like-minded individuals within each dataset. Practically, this partition of the data is then used as the axis along which different sets of claims are extracted and compared. This, we believe, also demonstrates the complementarity of our proposed claims extraction pipeline to Social Network Analysis.

\begin{figure*}[h!]
 \centering
 \makebox[\linewidth][c]{% 
 \begin{subfigure}{0.5\textwidth}
  \centering
  \includegraphics[width=0.7\linewidth]{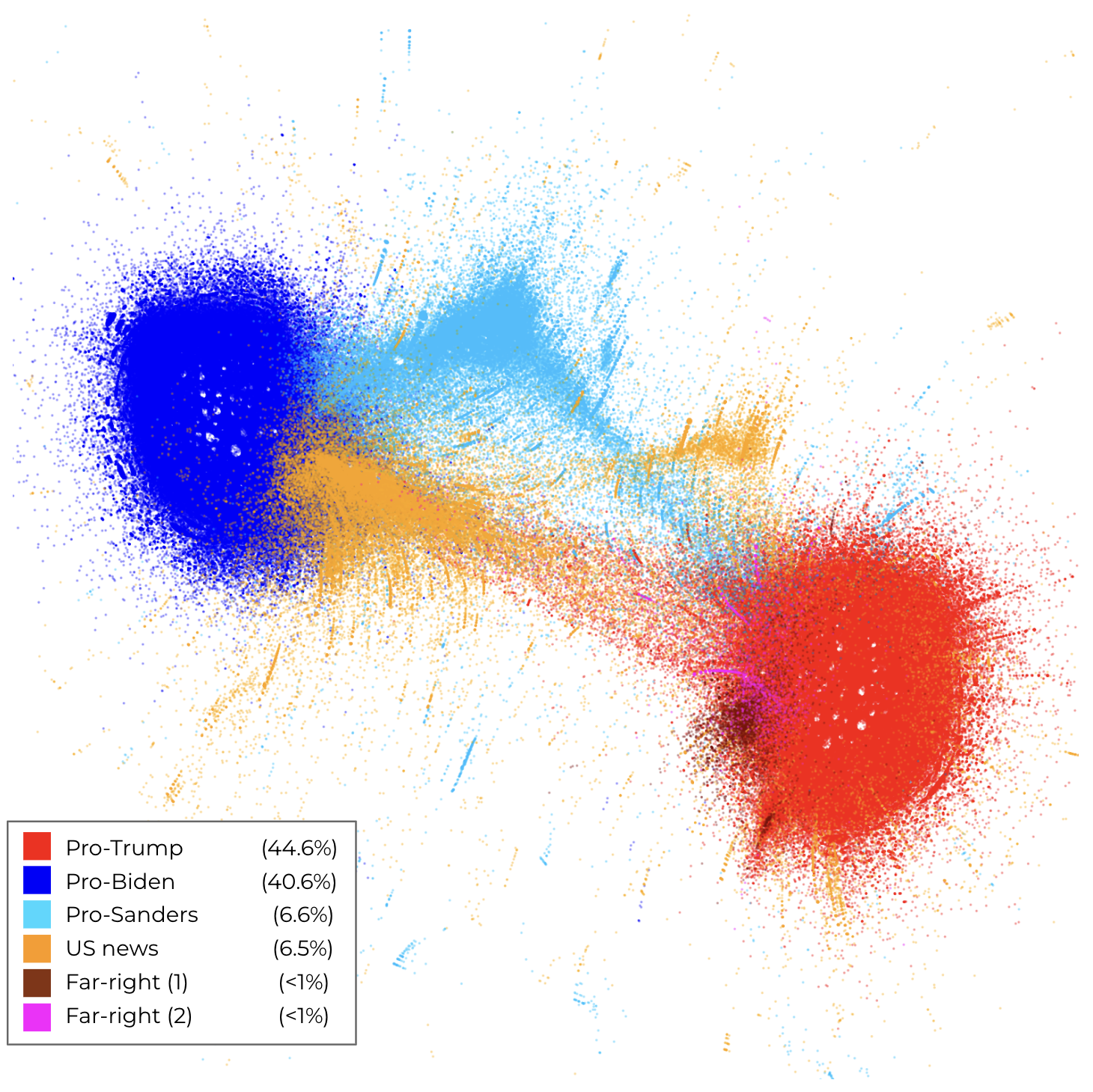}
  \caption{2020 US elections}
  \label{fig:sna_US}
 \end{subfigure}
 \begin{subfigure}{0.5\textwidth}
  \centering
  \includegraphics[width=0.7\linewidth]{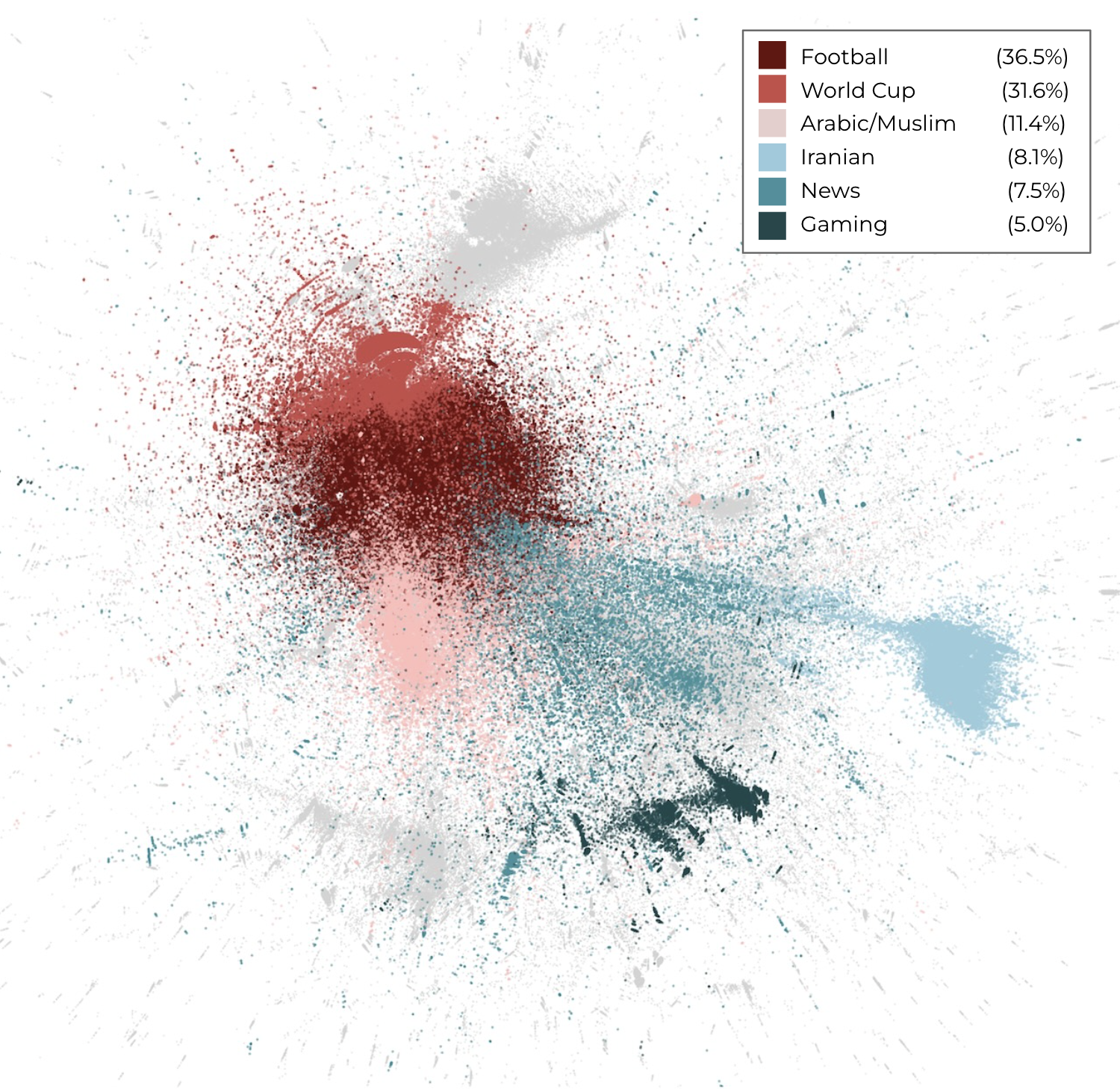}
  \caption{2022 FIFA World Cup}
  \label{fig:sna_FIFA}
 \end{subfigure}
 }
\caption{Retweet networks derived from the US elections and FIFA World Cup corpora respectively. Each point represents a Twitter users and users that are closer together retweet each other more. Colours correspond to communities, as derived using the Louvain community-detection algorithm.} 
\label{fig:sna}
\end{figure*}

\subsection{2020 US elections}

The first dataset used in this study was introduced by \citet{chausson_insight-inference_2026} and focuses on the 2020 US presidential election. The data was collected continuously from November 2019 to January 2023 using a range of keywords related to the election itself (e.g. ``2020 United States presidential election") or the names of various candidates (e.g. ``Donald Trump", ``Joe Biden", ``Bernie Sanders", ``Elizabeth Warren", etc.).

For our analysis of the dataset, we chose to restrict the dataset to tweets mentioning either ``Donald", ``Trump", ``Joe" or ``Biden” (case insensitive). This filtering was done using simple token matching. Secondly, we restricted the dataset to tweets published between the 8th of April 2020 (i.e. the day Bernie Sanders announced his withdrawal from the presidential race) and the 25th January 2021 (a few days after the presidential inauguration of Joe Biden on the 21st of January). This time window allows us to focus specifically on the stage of the election campaign that opposed Joe Biden to Donald Trump, while also covering the turbulent power transition between the Trump and Biden administrations (including, most significantly, the January 6 insurrection) and to capture delayed reactions to Joe Biden’s presidential inauguration. Finally, we used the language label returned by the Twitter API for each tweet to filter out all non-English tweets. These filtering steps reduced the size of the dataset to 71M unique tweets and 378M retweets.

The raw retweet network derived from the US elections corpus comprises 14M nodes and 128M weighted undirected edges, where each edge represents the number of times two accounts retweeted each other. Retaining only edges with a weight of at least 5 reduces this to a more manageable 1M nodes and 7.4M edges. We identify communities in this filtered network using the Louvain algorithm \cite{blondel-etal-2008-fast}, and characterize each community based on its 10 most-retweeted accounts by PageRank \cite{page-etal-1999-pagerank} and TF-IDF word clouds built from the self-authored Twitter bios of its members. This procedure surfaces six communities, three of which are central to this paper's analysis: 

\begin{itemize}
    \item a \textbf{Pro-Trump} community, organized around prominent Trump-supporting influencers (e.g. \@realDonaldTrump, \@charliekirk11, \@TrumpWarRoom), MAGA and Qanon rallying cries (``MAGA", ``KAG", ``WWG1WGA"), and religious and nationalist symbolism (``christian,'', ``God" ``patriot,'' ``American'');
    \item a \textbf{Pro-Biden} community, structured both around prominent Democratic accounts (\@JoeBiden) and prominent anti-Trump Republican accounts (\@ProjectLincoln), defined by opposition to Trump (``resist,'' ``resistance'') and a general commitment to Democratic values (``Democrat", ``liberal");
    \item a \textbf{Pro-Sanders} community, centered around the account of Bernie Sanders (\@BernieSanders) and with an explicitly progressive, policy-oriented self-identification (``socialist'', ``medicareforall''). 
\end{itemize}

The remaining three communities --- a mainstream-news cluster and two smaller far-right clusters --- are not used in this paper. Figure \ref{fig:sna_US} shows the resulting retweet network with each community, as visualised using the Force Atlas 2 algorithm \cite{jacomy_forceatlas2_2014}.

\subsection{2022 FIFA World Cup}

The 2022 FIFA World Cup dataset used in this paper was introduced by \citet{chausson_beyond_2026} and was collected from a wide range of English-language keywords related to the tournament. The dataset focuses exclusively on English-language tweets from UK accounts posted during the active phase of the World Cup: that is, from the $20^{th}$ of November (Opening Ceremony) to the $20^{th}$ of December 2022 (two days after the final). 

In their original paper, \citet{chausson_beyond_2026} already compute the retweet network from this dataset. Among the main UK-relevant sub-networks they identify we find 1) a ``Football" cluster, structured around UK football clubs, football leagues, and football news and statistics accounts, 2) a more international ``World Cup” cluster structured around the official accounts of the FIFA World Cup and various national teams, 3) an ``Arabic/Muslim" cluster of accounts with ties to Arab countries and/or Islam, and 4) a ``News" cluster composed of news media accounts and news aggregators (see Figure \ref{fig:sna_FIFA}). Their analysis also finds marked discursive differences between these sub-networks: the News and Football/World Cup clusters often framed the tournament as sport-washing, whereas the Arabic/Muslim cluster tended to reject such accusations as a racist double standard. Opposing interpretations of this kind are exactly what our claims extraction method should surface.

For the purpose of this study, we re-use the retweet network from \citet{chausson_beyond_2026} and focus specifically on the four clusters just listed. For simplicity, we merge the Football and World Cup clusters together into a single ``Football fandom" cluster. Overall, the version of the dataset we use contains tweets from 350K ``Football fandom" accounts, 59K Arabic/Muslim accounts and 39K News accounts.

\section{Method: LLM-Based Claims Extraction}
\label{sec:method}

\begin{figure*}[ht]
\centering
    \includegraphics[width=0.8\linewidth]{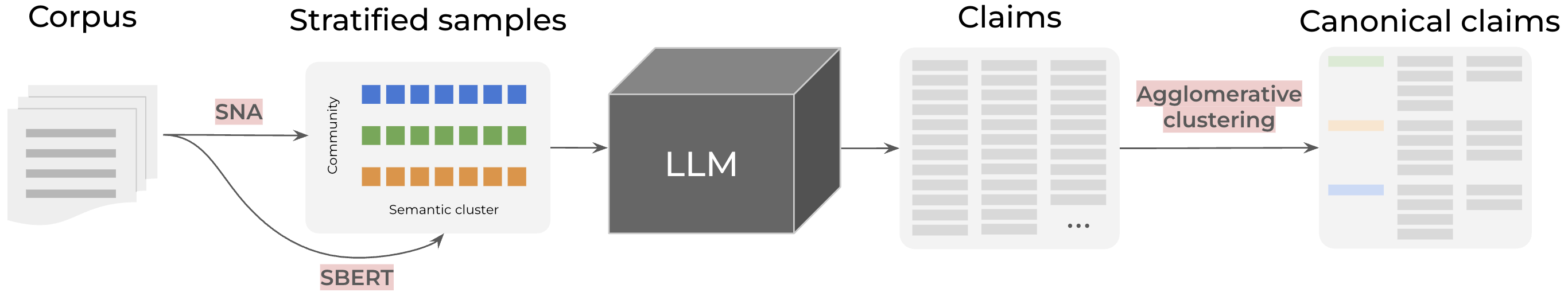}
    \caption{Overview of proposed claims extraction pipeline (``SNA" = Social Network Analysis).}
    \label{fig:pipeline}
\end{figure*}

Our pipeline extracts a catalogue of recurring claims from the corpus in three stages. First, we construct a sample of tweets by stratifying along \emph{semantic} and \emph{structural} dimensions, to ensure the sample represents the corpus's topical and community breadth. Second, we prompt an LLM to extract claims from each stratum independently. Third, because the same underlying claim is frequently extracted many times across the various strata, often with a different wording, we consolidate the raw LLM outputs into a catalogue of distinct claims through embedding-based clustering and manual review. An overall summary of this approach is shown in Figure \ref{fig:pipeline}. 

\subsubsection{Stratified Sampling}
\label{sec:stratified-sampling}

The first step of our pipeline consists in stratifying the data along \emph{structural} and \emph{semantic} dimensions, before sampling from the resulting strata. By ``structural" stratification, we essentially mean dividing the data according to the different retweet sub-networks surfaced by the community detection step presented earlier. Semantic stratification, on the other hand, consists in embedding the data with a Sentence-BERT model \cite{reimers2019sentence} (\texttt{all-mpnet-base-v2}), and applying K-means to partition the data into semantically similar clusters. This should be a relatively high number of clusters (a few hundreds), with the final number decided depending on how much granularity is wanted and how diverse the corpus is expected to be.

Stratification is done for analytical, technical and practical reasons. Analytically, insofar as a corpus is expected to reflect structural and ideological division (as both our US elections and FIFA corpora do), community-based stratification provides the foundation from which to compare across communities and therefore to generate insights into how communities differ from or resemble each other. 

From a technical perspective, community-based stratification will tend to coincide with stance when it comes to salient issues: \citet{aldayel2021stance} for instance demonstrate the importance of network-based features in their survey of stance detection methods. By sampling from each community independently, and from tweets that are semantically similar, the idea is therefore for the data within each sample to provide a shared discursive context that each datapoint individually might otherwise be missing. 

For instance, consider the tweet making the statement ``Trump only became racist when he won the election". While the actual meaning of this statement is ambiguous, it becomes a lot clearer when accompanied by the tweets ``Go ahead and find a picture of Donald Trump with a KKK member... I'll wait!" and ``Dear God... look at how racist Donald Trump is. Said nobody ever.". Together, these examples provide a clear pattern of sarcastic speech which, alongside more explicit statements (e.g. ``I believe @HerschelWalker when he says Donald Trump is not a racist.") and the fact that these tweets were posted by users from the Pro-Trump community, help clarify its intended meaning.

Finally, from a practical perspective, stratification can also help ensure that rare but distinct content remains represented down the line, in a way that random sampling does not. It can help limit the number of times the same claim is generated by the LLM with slightly different wording, as similar content will tend to be concentrated in the same batches.

In our worked example, given the large size of the US election dataset, we first randomly sub-sample 600K tweets for the rest of the analysis. This is not necessary for the FIFA World Cup dataset, which only contains around 574K tweets. Moreover, given that the US elections dataset spans a much longer period (10 months instead of 1), we stratify the US elections corpus into 600 clusters and the FIFA World cup corpus into 300\footnote{These steps essentially correspond to the ``hypersegment" phase of the ``hypersegment-and-merge" approach to Topic Modelling presented in \citet{chausson_beyond_2026}.}. 

Crossing semantic cluster with retweet-based communities defines 1,800 strata in total for the US elections corpus and 900 for the FIFA World Cup corpus. We randomly sample 30 tweets per stratum to constitute the smaller subset of data which will be fed into an LLM in the next stage of the pipeline: here, the number of 30 tweets per stratum is chosen on purely practical terms related to the LLM's context window size and the cost of API calls.

\subsection{LLM Prompting}
\label{sec:llm-prompting}

Each stratum's 30 tweets are passed to an LLM, along with a prompt instructing the model to identify the main claims expressed across the sample. In our case, we use Claude Sonnet 5: at the time of writing, Anthropic models rank the highest on text-to-text benchmarks \cite{arena_llm_2026}, and Claude Sonnet 5 is among the cheapest among their models. 

The full prompt, arrived at over several rounds of testing and revision, is given in the Appendix. Claims are defined following the definition given earlier: a declarative, single-clause statement that is evaluative or attributional. The prompt requires claims to be stripped of any explanation of how or why they hold, and in particular of any specific supporting event or piece of evidence. An event is treated as possible evidence for a claim, not as a claim itself, and the model is instructed to extract the underlying attribution the event would support rather than the event itself. Claims may read as ordinary statements of fact provided they take a side on a genuinely disputed question in the discourse under study (e.g. ``Joe Biden won the election"). Where relevant, hashtags and nicknames (e.g. \#StopTheSteal, \#ChinaVirus, ``GOAT") are treated as evidence for the claims they invoke. Finally, a single tweet may serve as an example for more than one claim and not every tweet need express a claim.

Applied across all 600 US elections strata, this yields 15,871 claims exemplified across 23,175 unique tweets. For the FIFA World Cup data, on the other hand, 3,693 claims are thus extracted, spread across 4,877 unique tweets.

\subsection{Claim Set Consolidation}
\label{sec:consolidation}

Because the same claim is frequently extracted independently across separate strata, often with differing wording, the raw output of the LLM prompting step requires substantial consolidation before it can be treated as a catalogue of distinct claims. To do this, we embed claims using a Sentence-BERT model (we use the \texttt{all-mpnet-base-v2} again) and then apply agglomerative clustering with average linkage to group the most similar claims together based on their cosine similarity. This requires a cosine similarity threshold to be defined: while there are different ways to obtain the threshold, we use the threshold-tuning procedure from  \cite{chausson_insight-inference_2026}, where pairs of claims are iteratively sampled and annotated as equivalent or not using the Probabilistic Bisection Algorithm. This allows us to find a threshold for each corpus and each community from a total of 197 binary annotations (i.e. around 33 annotations for each of the 6 thresholds). The agglomerative clustering step reduces the claims to 9,332 and 2,956 clusters for the US elections and FIFA corpora respectively\footnote{While this is only a modest decrease, many of the claims are singletons. The largest clusters typically contain more than 30 equivalent claims.}. 

Claim clusters supported by fewer than 5 tweets in total are set aside, on the grounds that claims recurring this rarely are unlikely to reflect a persistent claim of the kind this paper is interested in. For our example, this reduces the working set from 9,332 to 1,022 clusters of claims for the US elections corpus and from 2,956 to 442 for the FIFA World Cup corpus. Each of the remaining clusters is then reviewed manually: first, to correct residual clustering errors, by relocating misclassified claims, subdividing overly broad clusters, and merging duplicate clusters; and second, to relabel each cluster with a single representative claim, phrased at the level of abstraction its members actually support. This yields a final set of 594 ``canonical" claims for the US elections corpus, and 237 for the FIFA World Cup corpus.

\section{Evaluation}
\label{sec:evaluation}

The pipeline presented above is composed of three stages: i.e. Stratified Sampling, LLM prompting, and Claim Set Consolidation. The latter already includes a relatively thorough validation check as part of the pipeline, in the form of a manual review, revision and re-labelling of claim clusters. We therefore leave this stage aside in this evaluation to focus on the other two stages. 

In what follows, we first validate the accuracy of the LLM extraction step using a formal comparison between LLM-generated outputs and a sample of manually annotated data. Then, we validate the value of the Stratified Sampling step by performing an ablation study, where the outputs of the LLM-prompting step with and without stratification are compared.

\subsection{Evaluating LLM outputs}

We validate the LLM-prompting stage of our pipeline by focusing on the following two questions separately: 

\begin{itemize}
    \item \emph{Recall}: does the LLM extraction step capture the full set of claims each sample contains, according to an expert annotator? 
    \item \emph{Precision}: of all the claims the LLM extraction step identifies, what proportion is actually expressed in the data, according to an expert annotator? 
\end{itemize}

\subsubsection{Recall}
\label{sec:recall}

A sample of the data was manually annotated to measure the recall of the LLM-prompting step. Four different strata of 30 tweets were randomly selected per community from each dataset, for a total of 24 strata (i.e. 720 tweets). Table \ref{tab:strata} summarises the topical range of these various strata.

\begin{table}[h!]
\centering
\tiny
\begin{tabular}{clcl}
\hline
\multicolumn{2}{c}{\cellcolor[HTML]{C0C0C0}\textbf{US elections}} & \multicolumn{2}{c}{\cellcolor[HTML]{C0C0C0}\textbf{FIFA World Cup}} \\ \hline
 & Kamala Harris criticism &  & Qatar's global investments \\
 & Stolen election &  & Migrant workers deaths \\
 & Trump and racism &  & Portugal \\
\multirow{-4}{*}{\textbf{\begin{tabular}[c]{@{}c@{}}Pro-\\ Trump\end{tabular}}} & Biden vs Trump predictions & \multirow{-4}{*}{\textbf{\begin{tabular}[c]{@{}c@{}}Football\\ fandom\end{tabular}}} & Saudi Arabia \\ \hline
 & January 6th riots &  & Human rights and boycott \\
 & Biden mask mandates &  & Western media hypocrisy \\
 & Stock market performance &  & Fan experience \\
\multirow{-4}{*}{\textbf{\begin{tabular}[c]{@{}c@{}}Pro-\\ Biden\end{tabular}}} & Satanic accusations  & \multirow{-4}{*}{\textbf{\begin{tabular}[c]{@{}c@{}}Arabic/\\ Muslim\end{tabular}}} & Cristiano Ronaldo \\ \hline
 & Biden's competence &  & Grant Wahl death \\
 & Biden and racism &  & FIFA corruption \\
 & Critiques and support for Biden &  & General World Cup content \\
\multirow{-4}{*}{\textbf{\begin{tabular}[c]{@{}c@{}}Pro-\\ Sanders\end{tabular}}} & Black voters' support for Trump & \multirow{-4}{*}{\textbf{News}} & Morocco \\ \hline
\end{tabular}
\caption{Description of each stratum used for human annotation.}
\label{tab:strata}
\end{table}

Each stratum was then manually annotated by the first author, according to the same instructions provided to the LLM: i.e. to identify claims expressed in the sample and attribute them to relevant tweets, given the stated formal requirements (single-clause claims, which must be evaluative or attributional, stripped of supporting evidence). For each stratum, the first author performed a first pass over the data, extracting all the claims they could identify at the tweet level. They then reviewed the resulting list of claims, merging together paraphrased claims. 

These human-formulated claims were then independently validated by another (human) annotator. The latter was provided with 6 of the 24 strata (one per community per dataset) and asked, for each tweet/human-formulated claim pair, to label the claim as ``Correct" if the corresponding tweet did indeed make that claim, and ``Incorrect" otherwise. The second annotator labelled 124 out of 125 of tweet/human-formulated claim pairs as correct for the US elections strata ($99\%$), and 92 out of 97 for the FIFA strata ($95\%$).

A ``human" version of the claims thus created and validated, the LLM- and human-generated claims for each stratum were mapped against each other. That is, each human-generated claim was labelled as one of the following: ``Matched" if an exactly equivalent claim could be found among the LLM-generated set, ``Partial" if an imperfect match was found, and ``Missing" if no match was found. Moreover, ``Partial" matches were further annotated as follow: ``Generalisation" if the matched LLM-generated claim was more general than the human-formulated one (e.g. ``The media is biased against Qatar" vs ``The media is biased against the Qatar World Cup"); ``Specification" if it was more specific; and ``Related" if the two were equivalent in level of generality but differed in focus, emphasis or framing (e.g. ``Saudi Arabia’s win against Argentina is unexpected" vs ``Saudi Arabia's win over Argentina was one of the greatest upsets in World Cup history"). Table \ref{tab:eval} summarises this information for each dataset. 

\begin{table}[h!]
\centering
\tiny
\begin{tabular}{lcccc}
\cline{2-5}
 & \multicolumn{2}{c}{\textbf{FULL}} & \multicolumn{2}{c}{\textbf{ABLATION}} \\ \hline
 & \textbf{Elections} & \textbf{FIFA} & \textbf{Elections} & \textbf{FIFA} \\ \hline
\textbf{Recall} & \multicolumn{1}{l}{} & \multicolumn{1}{l}{} & \multicolumn{1}{l}{} & \multicolumn{1}{l}{} \\
\hspace{3mm}\textbf{Matched} & 0.29 & 0.23 & 0.27 & 0.20 \\
\hspace{3mm}\textbf{Partial} & 0.27 & 0.40 & 0.21 & 0.31 \\
\hspace{6mm}\textit{Generalisation} & 0.03 & 0.07 & 0.01 & 0.05 \\
\hspace{6mm}\textit{Specification} & 0.02 & 0.09 & 0.01 & 0.03 \\
\hspace{6mm}\textit{Related} & 0.22 & 0.24 & 0.19 & 0.23 \\
\hspace{3mm}\textbf{Missing} & 0.44 & 0.37 & 0.52 & 0.49 \\ \hline
\textbf{Precision} & 0.88 & 0.87 & 0.89 & 0.89 \\ \hline
\textbf{Nb of claims} & 165 & 163 & 199 & 181 \\ \hline
\end{tabular}
\caption{Evaluation of the ``full" claims extraction pipeline and the ``ablated" version of the pipeline on our manually annotated subset of US elections and FIFA World Cup Twitter data. ``Recall" measures the proportion of human-formulated claims also identified by the LLM and ``Precision" measures the proportion of tweet/LLM-generated claim pairs annotated as correct by a human annotator.}
\label{tab:eval}
\end{table}

Overall, 190 out of 338 of human-formulated claims in the US election sample has either a perfect or partial match in the LLM-generated ones ($56\%$). For the FIFA data, this figure stands at 132 out of 209 ($63\%$).

\subsubsection{Precision}
\label{sec:validation-precision}

To address the question of ``precision", a second set of annotations was created from the same 24 strata of data used earlier to measure recall. This time, the first author manually labelled all tweet/LLM-generated claim pairs to determine whether each claim was indeed expressed in the corresponding tweets (i.e. ``Correct" vs ``Incorrect"). A second annotator labelled the same 6 strata as before to measure agreement. Overall, the first author labelled $88\%$ of tweet/claim pairs from the US elections dataset as correct, and $87\%$ from the FIFA dataset. Inter-annotator reliability is imperfect but substantial, with a Cohen's Kappa agreement score of $0.75$ ($0.78$ for the FIFA dataset and $0.66$ for the US elections dataset). The mistakes the LLM tended to make can be divided into the following categories: 

\begin{itemize}
    \item \emph{Extrapolation}: Instead of focusing exclusively on claims that tweets explicitly make, the model will tend to surface implicit premises or consequences of those claims. For instance: e.g. the LLM extracting ``FIFA is corrupt" from ``FIFA is to be blamed for enabling situations like this". The great majority of errors fall into this category.
    \item \emph{Peripheral claims}: The model will sometimes not distinguish between the ``main" claim contained in the tweet (i.e. the point the tweet is intended to make) from more ``peripheral" claims: e.g. the claim ``Joe Biden won the election" in the tweet ``President-Elect Biden plans to introduce a mask mandate". While peripheral claims are not necessarily erroneous per se, and might actually be of analytical interest, it is not clear what leads the model to extract them in some cases but not others.
    \item \emph{Sarcasm and direct quotes}: The model will sometimes interpret sarcastic content literally, or attribute direct quotes to the author of the tweet.
    \item \emph{Meta-claims}: Very occasionally, the model will comment on the tweet's content or form rather than paraphrase its actual content: e.g. ``People are criticising Ronaldo" to describe a tweet critical of Ronaldo. 
\end{itemize}

Interestingly, we observe that none of the claims extracted by the LLM are actually fully hallucinated at the sample-level. That is, every claim the LLM identifies in the sample is actually correctly exemplified at least once.

\subsection{Evaluating stratification: Ablation study}

We evaluate the stratification and sampling step that precedes LLM-prompting, specifically in terms of its impact on the LLM's output.  We do this by running the LLM-prompting step on our subset of annotated data again but this time generate batches by randomly sampling tweets \emph{across} strata, such that tweets by different retweet communities and relating to different topics are mixed together. By then replicating the evaluation presented earlier, we are able to assess the impact of stratification on the pipeline. Going into this ablation study, we formulate the following hypotheses: 

\begin{itemize}
    \item \textbf{H1: Stratification improves recall}.  Semantic stratification should increase the likelihood of rare claims occurring repeatedly in the same batch, which could in turn help the model to ``catch" them.
    \item \textbf{H2: Stratification improves precision}. Homogeneous batches in terms of topic (semantic stratification) and stance (structural stratification) should help create a shared context, which in turn should help with the correct interpretation of individual tweets. 
    \item \textbf{H3: Stratification leads to fewer reformulations of the same claim}. Tweets making similar claims will be concentrated within fewer batches, which should limit mismatched reformulations of the same claim across prompting queries. 
\end{itemize}

Results from the ablation provide support for hypotheses \textbf{H1} and \textbf{H3} (see Table \ref{tab:eval}). We see that a greater proportion of the human-formulated claims are also extracted by the full pipeline than by the ablated pipeline: $56\%$ vs $48\%$ for the US elections data, and $63\%$ vs $51\%$ for the FIFA data. Moreover, as hypothesised, the full pipeline generates fewer claims than the ablated one (165 vs 199 for the US elections data and 163 vs 181 for the FIFA data). 

However, the ablated pipeline marginally outperforms the full pipeline for both datasets in terms of precision, with $88\%$ vs $89\%$ of tweet/claim pairs annotated as correct for the US data, and $87\%$ vs $89\%$ for the FIFA data. In other words, our results do not support \textbf{H2}. Still, a manual inspection of results provides anecdotal evidence that stratification helps in terms of stance and named entity resolution (see examples given in the \textbf{Method} section). This implies stratification must worsen the LLM's performance in other ways.

We speculate that the issue comes down to stratification worsening the model's tendency to \emph{extrapolate}. As discussed earlier though, extrapolated claims of this kind don't seem to ever be fully hallucinated, and always occur at least once unambiguously in the sample of 30 tweets. The effects of extrapolation can therefore be mitigated by using the pipeline to \emph{surface} claims rather than to measure their exact prevalence in the data.

\section{Claims catalogues}
\label{sec:findings}

To illustrate the capabilities of the pipeline, we here provide a snapshot of the claims catalogue extracted from each corpus. Tables \ref{tab:claims_US} and \ref{tab:claims_FIFA} show the top 20 most prevalent claims per community for each corpus. 

\begin{table*}[h!]
\centering
\scriptsize
\begin{tabular}{lclclc}
\hline
\multicolumn{2}{c}{{\color[HTML]{000000} \textbf{Pro-Trump}}} & \multicolumn{2}{c}{\textbf{Pro-Biden}} & \multicolumn{2}{c}{\textbf{Pro-Sanders}} \\ \hline
{\color[HTML]{000000} The 2020 election is stolen or rigged} & \cellcolor[HTML]{FF0000}{\color[HTML]{000000} 886} & Biden won the election & \cellcolor[HTML]{0000FF}{\color[HTML]{000000} 724} & Joe Biden is cognitively impaired & \cellcolor[HTML]{00D2FF}{\color[HTML]{000000} 373} \\
{\color[HTML]{000000} Joe Biden is cognitively impaired} & \cellcolor[HTML]{FD1C1C}{\color[HTML]{000000} 825} & Trump is responsible for January 6 & \cellcolor[HTML]{B6C7F4}{\color[HTML]{000000} 377} & Joe Biden is racist & \cellcolor[HTML]{8FF2FF}{\color[HTML]{000000} 236} \\
{\color[HTML]{000000} The media is biased or corrupt} & \cellcolor[HTML]{F78585}{\color[HTML]{000000} 592} & Trump is trying to cheat in the election & \cellcolor[HTML]{C1D2F3}{\color[HTML]{000000} 357} & Joe Biden is corrupt & \cellcolor[HTML]{BDFDFF}{\color[HTML]{000000} 192} \\
{\color[HTML]{000000} Joe Biden is corrupt} & \cellcolor[HTML]{F69797}{\color[HTML]{000000} 552} & Trump is a threat to democracy & \cellcolor[HTML]{CCDEF3}{\color[HTML]{000000} 336} & Joe Biden is a liar & \cellcolor[HTML]{C5FFFF}{\color[HTML]{000000} 184} \\
{\color[HTML]{000000} Trump is a good president} & \cellcolor[HTML]{F6D1D1}{\color[HTML]{000000} 403} & Donald Trump is a liar & \cellcolor[HTML]{D0E3F4}{\color[HTML]{000000} 327} & Joe Biden is a sexual predator or rapist & \cellcolor[HTML]{CAFFFF}{\color[HTML]{000000} 179} \\
{\color[HTML]{000000} Trump won the election} & \cellcolor[HTML]{F7D9D9}{\color[HTML]{000000} 353} & Donald Trump is corrupt & \cellcolor[HTML]{DFECF7}{\color[HTML]{000000} 263} & The media is biased or corrupt & \cellcolor[HTML]{D5FFFF}{\color[HTML]{000000} 154} \\
{\color[HTML]{000000} Joe Biden is illegitimate} & \cellcolor[HTML]{F8DCDC}{\color[HTML]{000000} 332} & Trump is a criminal & \cellcolor[HTML]{E3EEF8}{\color[HTML]{000000} 244} & Donald Trump is a liar & \cellcolor[HTML]{E4FFFF}{\color[HTML]{000000} 120} \\
{\color[HTML]{000000} Joe Biden is racist} & \cellcolor[HTML]{F9E3E3}{\color[HTML]{000000} 289} & Trump mishandled the COVID-19 pandemic & \cellcolor[HTML]{E7F1F9}{\color[HTML]{000000} 227} & Joe Biden is beholden to corporate interests & \cellcolor[HTML]{E7FFFF}{\color[HTML]{000000} 115} \\
{\color[HTML]{000000} Joe Biden is a liar} & \cellcolor[HTML]{FAE5E5}{\color[HTML]{000000} 276} & Donald Trump is racist & \cellcolor[HTML]{E8F1FA}{\color[HTML]{000000} 225} & Donald Trump is racist & \cellcolor[HTML]{EAFFFF}{\color[HTML]{000000} 107} \\
{\color[HTML]{000000} People support Donald Trump} & \cellcolor[HTML]{FAE6E6}{\color[HTML]{000000} 270} & Trump should be removed from office & \cellcolor[HTML]{EAF3FA}{\color[HTML]{000000} 215} & Joe Biden will lose to Trump & \cellcolor[HTML]{EBFFFF}{\color[HTML]{000000} 104} \\
{\color[HTML]{000000} Joe Biden is unpopular} & \cellcolor[HTML]{FBECEC}{\color[HTML]{000000} 230} & Trump should be held accountable or prosecuted & \cellcolor[HTML]{F0F6FC}{\color[HTML]{000000} 190} & The primary was rigged against Bernie & \cellcolor[HTML]{F7FFFF}{\color[HTML]{000000} 78} \\
{\color[HTML]{000000} Joe Biden is aligned with/controlled by China} & \cellcolor[HTML]{FCEFEF}{\color[HTML]{000000} 208} & Republicans are complicit in Trump's wrongdoing & \cellcolor[HTML]{F2F7FC}{\color[HTML]{000000} 182} & Biden won the election & \cellcolor[HTML]{F7FFFF}{\color[HTML]{FFFFFF} 77} \\
{\color[HTML]{000000} Joe Biden is controlled by others} & \cellcolor[HTML]{FDF5F5}{\color[HTML]{000000} 174} & Joe Biden will be a good president & \cellcolor[HTML]{F5F9FD}{\color[HTML]{000000} 169} & Joe Biden is a warmonger & \cellcolor[HTML]{F8FFFF}{\color[HTML]{000000} 75} \\
{\color[HTML]{000000} Joe Biden will be a bad president} & \cellcolor[HTML]{FEF7F7}{\color[HTML]{000000} 161} & Trump is aligned with/controlled by Russia & \cellcolor[HTML]{F5F9FD}{\color[HTML]{000000} 167} & Donald Trump is a fascist/nazi/dictator & \cellcolor[HTML]{FAFFFF}{\color[HTML]{000000} 72} \\
{\color[HTML]{000000} Trump will win the election} & \cellcolor[HTML]{FEF7F7}{\color[HTML]{000000} 159} & Trump lost the election & \cellcolor[HTML]{F7FAFD}{\color[HTML]{000000} 159} & Joe Biden is unpopular & \cellcolor[HTML]{FAFFFF}{\color[HTML]{000000} 72} \\
{\color[HTML]{000000} Joe Biden is harming America} & \cellcolor[HTML]{FEF8F8}{\color[HTML]{000000} 154} & Trump attempted an insurrection & \cellcolor[HTML]{FAFCFE}{\color[HTML]{000000} 146} & Joe Biden is not a true progressive & \cellcolor[HTML]{FAFFFF}{\color[HTML]{000000} 71} \\
{\color[HTML]{000000} Joe Biden is incompetent} & \cellcolor[HTML]{FEFAFA}{\color[HTML]{000000} 143} & Trump is a bad president & \cellcolor[HTML]{FBFDFE}{\color[HTML]{000000} 142} & Trump is a threat to democracy & \cellcolor[HTML]{FCFFFF}{\color[HTML]{000000} 66} \\
{\color[HTML]{000000} The Biden family is corrupt} & \cellcolor[HTML]{FEFBFB}{\color[HTML]{000000} 136} & Biden will win the election & \cellcolor[HTML]{FDFEFF}{\color[HTML]{000000} 134} & Trump mishandled the COVID pandemic & \cellcolor[HTML]{FCFFFF}{\color[HTML]{000000} 66} \\
{\color[HTML]{000000} Joe Biden is mentally weak} & \cellcolor[HTML]{FFFCFC}{\color[HTML]{000000} 129} & Trump is responsible for Covid deaths & \cellcolor[HTML]{FEFEFF}{\color[HTML]{000000} 131} & Joe Biden is unfit for office & \cellcolor[HTML]{FFFFFF}{\color[HTML]{000000} 59} \\
{\color[HTML]{000000} Big Tech is biased against Trump/conservatives} & \cellcolor[HTML]{FFFFFF}{\color[HTML]{000000} 105} & Joe Biden is a good person & \cellcolor[HTML]{FFFFFF}{\color[HTML]{000000} 123} & Donald Trump is corrupt & \cellcolor[HTML]{FFFFFF}{\color[HTML]{000000} 59} \\ \hline
\end{tabular}
\caption{Top 20 most common claims per retweet community in the US elections dataset.}
\label{tab:claims_US}
\end{table*}

\begin{table*}[h!]
\centering
\scriptsize
\begin{tabular}{lclclc}
\hline
\multicolumn{2}{c}{\textbf{Football fandom}} & \multicolumn{2}{c}{\textbf{Arabic/Muslim}} & \multicolumn{2}{c}{\textbf{News}} \\ \hline
{\color[HTML]{000000} Messi is a legendary player} & \cellcolor[HTML]{C5695E}126 & Messi is a legendary player & \cellcolor[HTML]{E75C42}{\color[HTML]{000000} 118} & Messi is a legendary player & \cellcolor[HTML]{4397B1}{\color[HTML]{000000} 99} \\
{\color[HTML]{000000} Jungkook's song ``Dreamers" is great} & \cellcolor[HTML]{D1877E}104 & Critics of Qatar are hypocrites & \cellcolor[HTML]{F3AB9D}{\color[HTML]{000000} 69} & Qatar is guilty of human rights violations & \cellcolor[HTML]{64A9BF}{\color[HTML]{000000} 85} \\
{\color[HTML]{000000} The World Cup is rigged} & \cellcolor[HTML]{EBC9C5}55 & Israel is unjustly imprisoning Ahmad Manasra & \cellcolor[HTML]{F5B8AC}{\color[HTML]{000000} 61} & FIFA is corrupt & \cellcolor[HTML]{7BB6C8}{\color[HTML]{000000} 75} \\
{\color[HTML]{000000} Morocco success is historic} & \cellcolor[HTML]{F5E4E2}35 & The media is biased against Qatar & \cellcolor[HTML]{F5BBB0}{\color[HTML]{000000} 59} & Migrant workers died for this World Cup  & \cellcolor[HTML]{A4CDD9}{\color[HTML]{000000} 57} \\
{\color[HTML]{000000} Messi/Argentina deserves the World Cup win} & \cellcolor[HTML]{F6E7E5}33 & The World Cup is rigged & \cellcolor[HTML]{F7C4BB}{\color[HTML]{000000} 53} & This World Cup final was the best ever & \cellcolor[HTML]{BDDBE4}{\color[HTML]{000000} 46} \\
{\color[HTML]{000000} Qatar hosted a fantastic World Cup} & \cellcolor[HTML]{F7EAE8}31 & Morocco success is historic & \cellcolor[HTML]{F9D4CD}{\color[HTML]{000000} 43} & Morocco's World Cup run is historic & \cellcolor[HTML]{CDE4EB}{\color[HTML]{000000} 39} \\
{\color[HTML]{000000} People know nothing about football} & \cellcolor[HTML]{F8EBEA}30 & The USA should be expelled & \cellcolor[HTML]{FAD9D3}{\color[HTML]{000000} 40} & Qatar persecutes LGBTQ people & \cellcolor[HTML]{D9EAEF}{\color[HTML]{000000} 34} \\
{\color[HTML]{000000} Western media is hypocritical in criticising Qatar} & \cellcolor[HTML]{F8EDEB}29 & Qatar hosted a successful World Cup & \cellcolor[HTML]{FADBD5}{\color[HTML]{000000} 39} & Politics should stay out of football & \cellcolor[HTML]{DBEBF0}{\color[HTML]{000000} 33} \\
{\color[HTML]{000000} It's football, not soccer} & \cellcolor[HTML]{F8EDEB}29 & Ronaldo is a football legend & \cellcolor[HTML]{FBE0DA}{\color[HTML]{000000} 36} & Qatar exploits migrant workers & \cellcolor[HTML]{E6F2F5}{\color[HTML]{000000} 28} \\
{\color[HTML]{000000} This World Cup is unpredictable} & \cellcolor[HTML]{F9EEED}28 & Messi deserved to win the World Cup & \cellcolor[HTML]{FCE9E6}{\color[HTML]{000000} 30} & West criticism of Qatar is hypocritical & \cellcolor[HTML]{EBF4F7}{\color[HTML]{000000} 26} \\
{\color[HTML]{000000} FIFA is corrupt} & \cellcolor[HTML]{FAF1F0}26 & This World Cup is the best ever & \cellcolor[HTML]{FDECE9}{\color[HTML]{000000} 28} & Messi is central to Argentina's success & \cellcolor[HTML]{EBF4F7}{\color[HTML]{000000} 26} \\
{\color[HTML]{000000} Qatar is guilty of human rights violations} & \cellcolor[HTML]{FAF1F0}26 & FIFA is corrupt & \cellcolor[HTML]{FDF0ED}{\color[HTML]{000000} 26} & Qatar's national team is bad & \cellcolor[HTML]{EFF7F9}{\color[HTML]{000000} 24} \\
{\color[HTML]{000000} Politics should stay out of football} & \cellcolor[HTML]{FBF5F4}23 & Politics should be kept out of football & \cellcolor[HTML]{FEF4F3}{\color[HTML]{000000} 23} & Moroccan fans are prone to violence & \cellcolor[HTML]{F2F8FA}{\color[HTML]{000000} 23} \\
{\color[HTML]{000000} The 2022 World Cup final was one of the best ever} & \cellcolor[HTML]{FCF6F5}22 & It's football, not soccer & \cellcolor[HTML]{FEF4F3}{\color[HTML]{000000} 23} & FIFA is complicit in human rights abuses & \cellcolor[HTML]{F9FCFD}{\color[HTML]{000000} 20} \\
{\color[HTML]{000000} African teams are achieving historic success} & \cellcolor[HTML]{FDF9F8}20 & Qatar played badly as host team & \cellcolor[HTML]{FEF6F4}{\color[HTML]{000000} 22} & It's football, not soccer & \cellcolor[HTML]{FBFDFE}{\color[HTML]{000000} 19} \\
{\color[HTML]{000000} Women referees are a historic milestone} & \cellcolor[HTML]{FDF9F8}20 & The USA insulted Iran & \cellcolor[HTML]{FFFCFC}{\color[HTML]{000000} 18} & VAR is corrupt or biased & \cellcolor[HTML]{FBFDFE}{\color[HTML]{000000} 19} \\
{\color[HTML]{000000} Qatar is the first host nation to be eliminated} & \cellcolor[HTML]{FDFAFA}19 & Israel is committing human rights abuses & \cellcolor[HTML]{FFFCFC}{\color[HTML]{000000} 18} & The World Cup is rigged & \cellcolor[HTML]{FBFDFE}{\color[HTML]{000000} 19} \\
{\color[HTML]{000000} Argentina will win the World Cup} & \cellcolor[HTML]{FFFFFF}15 & Women referees are a historic milestone & \cellcolor[HTML]{FFFCFC}{\color[HTML]{000000} 18} & Argentina is the favourite to win & \cellcolor[HTML]{FDFEFF}{\color[HTML]{000000} 18} \\
{\color[HTML]{000000} The opening ceremony was excellent} & \cellcolor[HTML]{FFFFFF}15 & Football fans should support Palestine & \cellcolor[HTML]{FFFCFC}{\color[HTML]{000000} 18} & Messi deserved to win the World Cup & \cellcolor[HTML]{FFFFFF}{\color[HTML]{000000} 17} \\
{\color[HTML]{000000} FIFA banned the One Love armband} & \cellcolor[HTML]{FFFFFF}15 & Qatar is the victim of a double standard & \cellcolor[HTML]{FFFFFF}{\color[HTML]{000000} 16} & The Iranian team bravely protested & \cellcolor[HTML]{FFFFFF}{\color[HTML]{000000} 17} \\ \hline
\end{tabular}
\caption{Top 20 most common claims per retweet community in the FIFA World Cup dataset.}
\label{tab:claims_FIFA}
\end{table*}

In the context of the US elections data, the claim that the 2020 election was stolen dominates the Pro-Trump community. The Pro-Biden community, on the other hand, depicts Trump as an authoritarian leader, a cheat, or more generally as a threat to democracy. These claims mirror each other, each providing insights into how events like the election results, January 6th riots and Inauguration Day were interpreted within each community. 

Claims that circulate within the Pro-Trump and Pro-Biden communities tend to be relatively homogeneous. Each community's claims unanimously criticise the opposite party's candidate while defending its own. Rather than discussing campaign promises or policies, these claims mostly pertain to the character of each candidate: Biden and Trump are described as corrupt, lying, sexually depraved, racist, mentally challenged or incompetent. Accusations of racism against Biden in the Pro-Trump community are particularly interesting and might reflect a contested Black voter base. 

The Pro-Sanders community contrasts somewhat with the other two in that, with Bernie Sanders out of the presidential race, the community's top claims are exclusively critical and target both Trump and Biden. While Biden appears to be the main target, this might not so much reflect a ``pro-Trump" stance than a more animated conversation around Biden, as the Democratic nominee and therefore the (contested) representative of this electorate. Some of the criticism formulated by Pro-Sanders accounts is also unique to that community. For instance, it paints Biden as a warmonger, as ``not a true progressive", and as a corporate crony.

Claims extracted from the FIFA dataset reflect less polarisation. The claim that Messi is a legendary player was the most prevalent across all three retweet communities (Football fandom, Arabic/Muslim cluster and News). Moreover, claims that the World Cup is rigged, that Morocco's win was historic, that FIFA is corrupt, that the competition was a success, or that politics should be kept out of football circulated within all communities, albeit at different rates. 

Still, some differences are noticeable. The Arabic/Muslim community for instance criticised Israel for its crimes in Palestine, criticised Western media for what it perceived as a hypocritical and racist treatment of Qatar, and generally defended Qatar more adamantly than any of the other communities. Conversely, accusations against Qatar --- especially its treatment of LGBTQ minorities and migrant workers --- were more prevalent in the News cluster.

\section{Discussion}\label{sec:discussion}

Large language models are now capable enough that their appeal for qualitative, interpretive tasks is obvious \cite{bail2024generative}, and they are increasingly deployed through agentic systems that execute multi-step analyses with little direct supervision. Underneath any such system, however, is still a pipeline: a sequence of decisions about how a corpus is sampled, how the model is prompted, and how its raw output is consolidated. The methodological task for CSS is to specify what that pipeline should look like, where it fails, and when its output can be trusted for downstream analysis.

This is the gap this paper tries to fill: formalising inductive, LLM-assisted discourse analysis using claims as the unit of analysis, proposing a concrete pipeline for identifying claims in social media data at scale, and subjecting that pipeline's output to systematic evaluation. We see the validation presented in this paper as a step toward the kind of evidentiary foundation this broader shift --- towards LLMs and agentic systems as tools for social science research at scale --- will need if it is to be trusted \cite{pangakis2023validation, tornberg2024best}.

Beyond the specific pipeline, this paper argues for inductive claims extraction as a task in its own right. A claims catalogue combines what topic modelling and stance detection each provide on their own --- the inductive, corpus-derived structure of the former and the positional content of the latter --- at a granularity neither reaches. This does not make claims a replacement for topics or stances, which remain cheaper to obtain and require no LLM validation. Rather, claims are a finer unit of analysis particularly suited for questions about what communities actually assert, with possible applications to cultural sociology and research on structural phenomena like polarisation or echo chambers.

Stratification is the central design choice of the pipeline: sampling within retweet community and semantic cluster is meant to give the LLM the shared context needed to interpret individual tweets, to protect rare content, and to limit redundant paraphrasing of the same claim. Our evaluation suggests this design mostly delivers on that promise. Precision, as measured against human annotations, is high and stable. More reassuringly still, none of the claims extracted from any given sample were found to be entirely unsupported. Recall is comparatively weaker: 56\% and 63\% of the claims an expert annotator formulated by hand were also recovered by the pipeline in the tested samples, either exactly or in a related form. The ablation study, in which the same LLM-prompting step is run on randomly mixed, unstratified batches, supports two of our three hypotheses about why stratification should matter. Stratification improves recall substantially, from 48\% to 56\% for the US elections data and from 51\% to 63\% for the FIFA data, and produces a smaller, less redundant catalogue of claims. Stratification does not, however, improve precision: if anything, the ablated pipeline is marginally more precise.

We read this as a genuine, if modest, trade-off. The precision cost of stratification appears to come from the same mechanism that improves its recall: by supplying the model with more context, stratification makes it more prone to \emph{extrapolation} (our largest single error category), where the model surfaces an implicit premise or consequence of a claim rather than the claim a tweet actually makes. Notably, this is the very behaviour that \citet{hoyle2023decompositions} deliberately exploit to elicit implicit propositions: whether it counts as error or insight depends on whether the analyst wants what a text says or what it presupposes. This trade-off can inform how our proposed pipeline ought to be used. Extrapolation degrades the pipeline's reliability as a precise measure of claim prevalence, an error that would propagate into any downstream analysis built on those counts \cite{egami2023surrogates, baumann2025hacking}. However, since no claims in our evaluation were found to be entirely hallucinated on the sample level, extrapolation does not compromise the pipeline's reliability if used simply to surface (and potentially rank) which claims exist in a corpus. This does not rule out prevalence estimation altogether: a claims-extraction pipeline of the kind presented here pairs naturally with more ``top-down" claim-matching methods discussed in the \textbf{Related work} section.

Still, our analysis raises concerns on the question of recall. Not only is recall lower than precision on average in our evaluation, it is also the harder of the two failure modes to detect in practice. After all, missed claims are, by definition, invisible in the pipeline's outputs rather than present and flagged for review. This can have real consequences for downstream analysis, especially if claims tend to be \emph{systematically} rather than randomly under-detected. This takes us to the issue of bias. An LLM's generation behaviour is known to be shaped by the prevalence of a given framing in its training data \cite{mccoy2024embers, kandpal2023longtail, feng2023pretraining, santurkar2023whose}. As such, claims that run counter to a dominant narrative may be both under-detected and more prone to being flattened or distorted by extrapolation than claims that echo familiar, well-rehearsed framings. Our own evaluation was not designed to measure bias in this way. However researchers using this pipeline (or others like it) on discourse involving minority or counter-narrative viewpoints should treat this as an open risk and take explicit precautions against the pipeline quietly caricaturing the positions it is least equipped to recognise.

 \section{Limitations and future work}

 \textbf{Batch size}. We evaluate a single sample size of 30 tweets per stratum, chosen on practical grounds related to context-window size and API cost. Larger batches would supply more context, and therefore more opportunity both for reconstructing essential context and for extrapolation. More work therefore needs to be done to establish the net effect of batch size as an independent factor.

\textbf{Model choice}. We evaluate a single LLM (Claude Sonnet 5). We do not attempt a comparison across model families, both because such a comparison would likely be obsolete within a short time and because general-purpose LLM benchmarks already track this kind of progress. What would be more durable, and what we see as an open question for future work, is how fundamental architectural choices bear on a task like claims extraction specifically, and on the failure modes this paper identifies.

\textbf{Domain generality}. We explicitly scope our evaluation to the study of \emph{political} discourse online and choose our corpora accordingly. It remains an open question how the pipeline would perform on discourse that is more thoroughly non-political, and whether the same error patterns would still dominate.

\textbf{Prompt sensitivity}. We did not systematically evaluate the impact of alternative prompt formulations on the final results. Prompt design is known to have a substantial effect on LLM outputs \cite{barrie2024promptstability, baumann2025hacking}. A more systematic evaluation of its impact in the context of our pipeline would be valuable.

\textbf{Dominant versus counter-narrative performance.} We suspect that issues of bias in this pipeline intersect specifically with how well it performs on dominant narratives vs more marginal or counter-hegemonic discourse. Future work, particularly work explicitly concerned with bias, should directly test this, for instance, by deliberately sampling and evaluating on discourse from minority communities across several domains.

\textbf{A single ground-truth annotator.} The claims used as ground truth throughout our evaluation were produced by a single annotator: the first author, who has worked extensively with this data and can reasonably be considered an expert on it. A second annotator validated a subset of this ground truth at high agreement (99\% for the US elections strata, 95\% for the FIFA strata). Still, this does not tell us whether a different expert would have surfaced a meaningfully different set of claims. Future work should scale this kind of evaluation up to several trained annotators.
 
\section{Conclusion}

This paper formalised claims as a unit of analysis for the inductive study of political discourse, and proposed and validated an LLM-based pipeline for surfacing them at the scale social media corpora require. Applied to two thematically distinct Twitter corpora, the pipeline distils hundreds of thousands of tweets into reviewed catalogues of 594 and 237 claims (released in full as supplementary material), with high precision, moderate recall, and a measurable benefit from stratifying both structurally (by retweet community) and semantically (by embedding-based cluster). As LLM-driven systems are increasingly deployed for this kind of inductive discourse analysis, we hope the pipeline and its validation offer a template Computational Social Scientists can build on and interrogate.
 
\bibliography{references}

\appendix

\section{LLM prompt}
\label{app:prompt}

\vspace{5mm}

Below is the prompt used during the LLM-prompting step of the pipeline on the US elections dataset:

\vspace{5mm}

\begin{small}
\begin{verbatim}
CONTEXT: Here are 30 tweets from a dataset 
related to the 2020 US presidential election, 
indexed 0–29. The objective is to identify 
the main claims being made across the sample.

TASK: Identify the main claims expressed across
the sample as a whole, with the indices of 
supporting tweets. 

DEFINITION: A claim is a declarative statement, 
simple enough to be expressed as a sentence 
with a single clause. We are specifically 
interested in claims that make a point: i.e. 
statements that take a side on a disputed 
question, or that attribute a character, 
quality, motive, trajectory, or allegiance to 
a person, group, institution, or abstract 
entity (whether positively or negatively). 
E.g. "Trump is a threat to democracy", 
"Trump is a great business man", "Joe Biden
is a corporate crony", "Joe Biden is a good
leader", "the mainstream media is biased",
"China is a threat".

This includes claims that read as factual but 
make a point in context: e.g. "Joe Biden won 
the election" counts as a claim, because it
takes a side on a disputed question, directly
opposed by the rival claim "the election was
stolen".

This excludes:
- Statements that are genuinely uncontested 
and  make no evaluative point in this 
discourse. E.g. "early voting starts today",
"the debate is on Thursday";
- Statements that report a specific event or
action, even a widely-discussed one (e.g. 
"Biden signed an executive order banning the
term 'China Virus'"). An event is not itself 
a claim, though it may be evidence for one. 
Extract the attribution the event is evidence 
for instead (e.g. "Biden is aligned with 
China," "Biden is restricting free speech").

CRITICAL: A claim must be expressed as a
SINGLE clause sentence: it asserts that
something is or happened, without explaining
how or why, and without providing evidence 
for why it's true or false. Phrase the claims
as simply as possible, avoiding unnecessary 
prepositions or compounds phrases. 

EXAMPLES:
- WRONG: "Bernie Sanders lost because the DNC 
rigged the primary against him" (unnecessary
compound sentence)
- CORRECT: "The DNC rigged the primary 
against Bernie 
Sanders" (--> rephrase as single clause 
sentence)

- WRONG: "China unleashed Covid onto America 
to harm Trump" (should not provide a "why")
- CORRECT: "China unleashed Covid onto 
America", "China wants to harm Trump" (-->
create two separate claims)

- WRONG: "The media is biased and dishonest" 
(unnecessary compound noun phrase)
- CORRECT: "The media is biased" (--> keep
the most relevant of the two adjectives)

- WRONG: "The Democrats are stealing the
election through mail-in ballot fraud"
(should not provide a "how")
- CORRECT: "The Democrats are stealing the
election" (--> drop unnecessary preposition)

- WRONG: "Biden's ban on the term 'China
virus' suppresses free speech" (still event-
anchored, even though grammatically single-
clause)
- CORRECT: "Biden is restricting free speech"
(fully abstracted attribution, event dropped
entirely)

NOTES: Several tweets can support the same
underlying claim even when they refer to
different events or evidence, or use
different wording. Not every tweet needs to
be included: some will express no claim of 
this kind at all. The same tweet can express
several claims. Only include a tweet as an
example if it genuinely supports the stated 
claim. Tweets may express claims indirectly
via hashtags or nicknames (e.g., 
"#StopTheSteal," "#BunkerBoy",
"#ChinaVirus"): these should be counted as
evidence of the claims they invoke.

FORMAT: Return the result as a list of 
dictionaries, each with keys "Claim" (string) 
and "Example_tweets" (list of integers). For 
instance:
[{"Claim": "Joe Biden is a corporate crony.", 
"Example_tweets": [0, 11, 14, 15, 27, 28]},
{"Claim": "China is hostile to Trump.", 
"Example_tweets": [2, 8, 13, 25]}]

CONSTRAINT: Return ONLY the resulting JSON. 
No explanation, no comments, no additional 
formatting.
\end{verbatim}
\end{small}

\vspace{5mm}

The following, on the other hand, is the prompt used for the FIFA World Cup dataset. The main differences is in the examples provided, which match the corpus on which the prompt is deployed: 

\vspace{5mm}

\begin{small}
\begin{verbatim}
CONTEXT: Here are 30 tweets from a dataset
related to the 2022 FIFA World Cup, indexed 0–
29. The objective is to identify the main 
claims being made across the sample.

TASK: Identify the main claims expressed 
across the sample as a whole, with the 
indices of supporting tweets. 

DEFINITION: A claim is a declarative 
statement, simple enough to be expressed as a
sentence with a single clause. We are 
specifically interested in claims that make a
point: i.e. statements that take a side on a 
disputed question, or that attribute a 
character, quality, motive, trajectory, or 
allegiance to a person, group, institution,
or abstract entity (whether positively or 
negatively). E.g. "The News is biased against
Qatar", "Qatar is guilty of human rights 
abuses", "Politics has no place in football".

This includes claims that read as factual but
make a point in context: e.g. "Qatar 
criminalises homosexuality".

This excludes:
- Statements that are genuinely uncontested
and make no evaluative point in this 
discourse. E.g. "The World Cup final is on 
the 18th of December", "The 2022 World Cup is
in Qatar";
- Statements that report a specific event or
action, even a widely-discussed one (e.g. 
"Argentina won the World Cup"). An event is
not itself a claim, though it may be evidence
for one. Extract the attribution the event is
evidence for instead, if there is one (e.g. 
"The World Cup was rigged in favour of 
Argentina").

CRITICAL: A claim must be expressed as a 
SINGLE clause sentence: it asserts that 
something is or happened, without explaining
how or why, and without providing evidence 
for why it's true or false. Phrase the claims
as simply as possible, avoiding unnecessary 
prepositions or compounds phrases. 

EXAMPLES:
- WRONG: "Argentina only won the World Cup 
because FIFA rigged the game in their favour"
(unnecessary compound sentence)
- CORRECT: "FIFA rigged the World Cup in 
favour of Argentina" (--> rephrase as single
clause sentence)

- WRONG: "The BBC boycotted the opening 
ceremony because it is biased against Qatar"
(should not provide a "why")
- CORRECT: "The BBC boycotted the opening 
ceremony", "The BBC is biased against Qatar"
(--> create two separate claims)

- WRONG: "The media is biased and dishonest"
(unnecessary compound noun phrase)
- CORRECT: "The media is biased" (--> keep 
the most relevant of the two adjectives)

- WRONG: "Qatar cheated to get the nomination
by bribing committee members" (should not 
provide a "how")
- CORRECT: "Qatar cheated to get the 
nomination" (--> drop unnecessary preposition)

- WRONG: "Qatar's ban on rainbow clothing 
suppresses free speech" (still event-
anchored, even though grammatically single-
clause)
- CORRECT: "Qatar is suppressing free speech"
(fully abstracted attribution, event dropped 
entirely)

NOTES: Several tweets can support the same 
underlying claim even when they refer to 
different events or evidence, or use 
different wording. Not every tweet needs to 
be included: some will express no claim of 
this kind at all. The same tweet can express
several claims. Only include a tweet as an 
example if it genuinely supports the stated 
claim. Tweets may express claims indirectly 
via hashtags or nicknames (e.g., 
"#BoycottQatar2022", 
"#HumanRightsViolations," 
"#IsraeliTerrorism"): these should be counted
as evidence of the claims they invoke.

FORMAT: Return the result as a list of 
dictionaries, each with keys "Claim" (string)
and "Example_tweets" (list of integers). For
instance:
[{"Claim": "The News is biased against 
Qatar.", "Example_tweets": [0, 11, 14, 15, 
27, 28]}, {"Claim": "Qatar is guilty of human
rights violations.", "Example_tweets": [2, 8,
13, 25]}]

CONSTRAINT: Return ONLY the resulting JSON.
No explanation, no comments, no additional 
formatting.        
\end{verbatim}
\end{small}

\end{document}